\documentclass[letterpaper, 10 pt, conference]{ieeeconf}  

\IEEEoverridecommandlockouts                              

\usepackage{graphicx}
\usepackage{color}
\usepackage{float}
\usepackage{amsmath}
\usepackage{amssymb}
\usepackage{algorithmic}
\usepackage{algorithm}
\usepackage{arydshln}
\usepackage{bm}
\usepackage{upgreek}
\usepackage{booktabs}
\usepackage{subcaption}
\usepackage{tikz}

\title{\LARGE \bf
Incremental Residual Reinforcement Learning Toward Real-World Learning for Social Navigation
}

\author{Haruto Nagahisa$^{1}$$^{\dagger}$, Kohei Matsumoto$^{2}$$^{\dagger}$, Yuki Tomita$^{1}$, Yuki Hyodo$^{1}$, and Ryo Kurazume$^{2}$
\thanks{$^{1}$Haruto Nagahisa, Yuki Tomita, and Yuki Hyodo are with the Graduate School of Information Science and Electrical Engineering, Kyushu University, Fukuoka, Japan {\tt\small nagahisa@irvs.ait.kyushu-u.ac.jp, tomita@irvs.ait.kyushu-u.ac.jp, hyodo@irvs.ait.kyushu-u.ac.jp}}%
\thanks{$^{2}$Kohei Matsumoto and Ryo Kurazume are with the Faculty of Information Science and Electrical Engineering, Kyushu University, Fukuoka, Japan {\tt\small matsumoto@ait.kyushu-u.ac.jp, kurazume@ait.kyushu-u.ac.jp}}%
\thanks{$^{\dagger}$Authors contributed equally}%
}

\begin{document}

\maketitle
\thispagestyle{empty}
\pagestyle{empty}

\begin{abstract}

As the demand for mobile robots continues to increase, social navigation has emerged as a critical task, driving active research into deep reinforcement learning (RL) approaches. However, because pedestrian dynamics and social conventions vary widely across different regions, simulations cannot easily encompass all possible real-world scenarios. Real-world RL, in which agents learn while operating directly in physical environments, presents a promising solution to this issue. Nevertheless, this approach faces significant challenges, particularly regarding constrained computational resources on edge devices and learning efficiency. In this study, we propose incremental residual RL (IRRL). This method integrates incremental learning, which is a lightweight process that operates without a replay buffer or batch updates, with residual RL, which enhances learning efficiency by training only on the residuals relative to a base policy. Through the simulation experiments, we demonstrated that, despite lacking a replay buffer, IRRL achieved performance comparable to those of conventional replay buffer-based methods and outperformed existing incremental learning approaches. Furthermore, the real-world experiments confirmed that IRRL can enable robots to effectively adapt to previously unseen environments through the real-world learning.

\end{abstract}

\section{INTRODUCTION}

In recent years, the demand for mobile robots has increased significantly across various fields, including cleaning, transportation, and guidance. Consequently, social navigation, which is the ability of robots to navigate dynamic environments populated by humans, has emerged as a critical research area. In social navigation, robots must adhere to implicit social norms while moving safely. Deep reinforcement learning (DRL) has accelerated the development of navigation policies that enable robots to achieve their goals while respecting social conventions \cite{c2}–\cite{c29}. However, because pedestrian behaviors and social norms vary widely across different locations, navigation systems must account for a vast array of potential scenarios \cite{c26}. Although many DRL methods are trained exclusively using simulations, it is inherently difficult to capture the full complexity of real-world situations through simulation alone \cite{c30}, \cite{c27}. Thus, adapting to unexpected changes in dynamic environments remains a significant challenge for DRL-based approaches.\par
To address this limitation, real-world reinforcement learning (RL), an approach in which agents learn directly while operating in physical environments, has emerged as a promising solution. However, real-world RL introduces its own set of challenges. Autonomous navigation requires the simultaneous execution of multiple resource-intensive processes, such as localization and object detection. Moreover, edge devices mounted on mobile robots typically face strict constraints on computational resources \cite{c9}, \cite{c10}. Furthermore, unlike in simulations, acquiring vast amounts of interaction data is impossible in the physical world, making learning efficiency paramount. Owing to these practical constraints, modern DRL architectures rely heavily on extensive replay buffers and batch updates to enhance performance \cite{c9}, \cite{c10}; therefore, they are rarely designed with the limitations of real-world, on-device training in mind. \par
In this study, we propose incremental residual RL (IRRL) to bridge this gap. IRRL integrates incremental learning, a lightweight training paradigm that eliminates replay buffers and batch updates, with residual RL \cite{c28}, \cite{c1}, which aims to improve learning efficiency by training only the residual actions relative to a predefined base policy. \\
The main contributions of this study are as follows: 

\begin{figure}[t]
  \centering
  \includegraphics[width=1.0\linewidth]{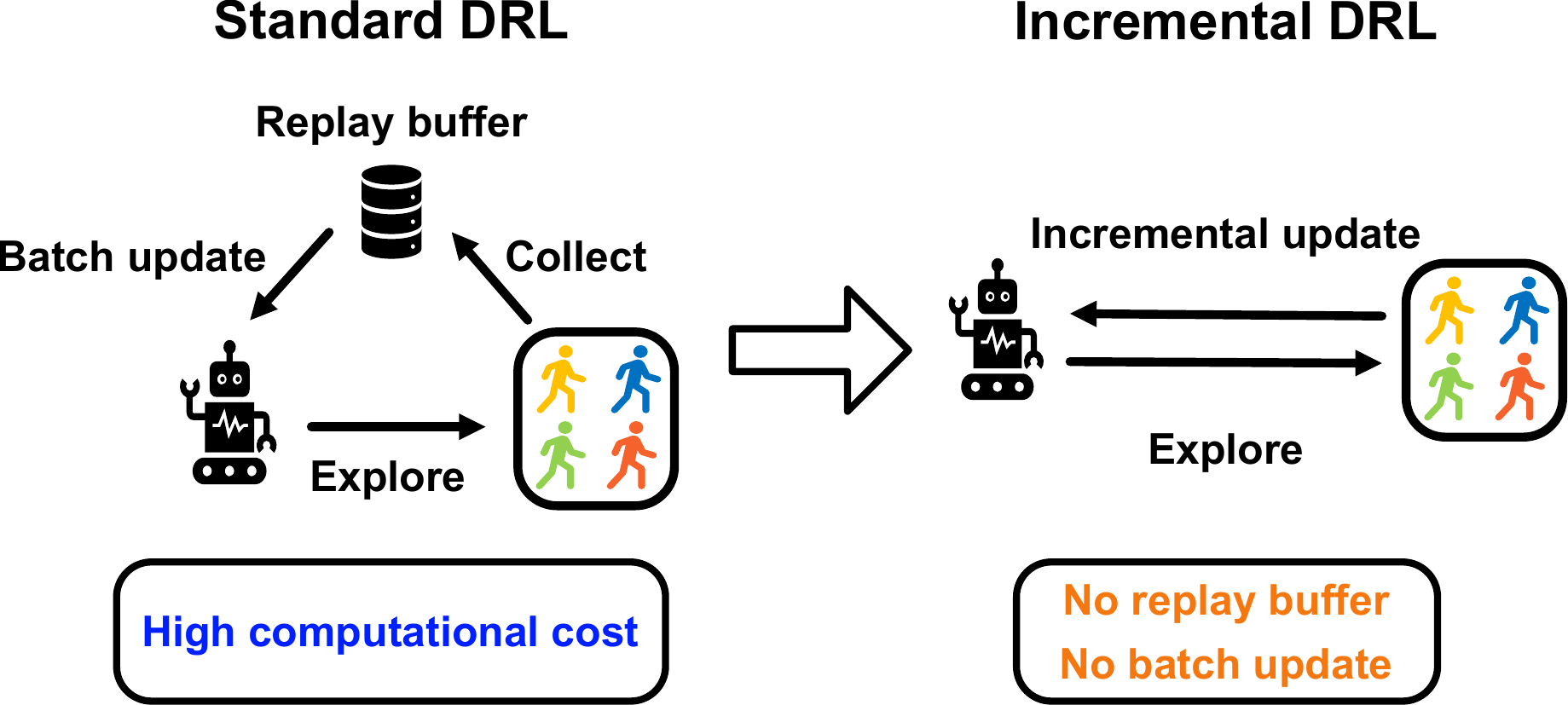}
  \caption{Overview of standard deep reinforcement learning (DRL) and incremental learning. In standard DRL, collected data is stored in a replay buffer and used for batch updates during training. However, this approach is computationally expensive in resource-constrained environments. In contrast, incremental learning trains the model using only the most recently collected data, making it particularly suitable for learning under strict resource constraints.}
  \label{fig:incremental_learning}
\end{figure}

\begin{itemize}

  \item IRRL is introduced, which is a novel framework that integrates incremental learning and residual RL and is specifically tailored for the real-world learning of social navigation. To the best of our knowledge, this is the first attempt to apply such an approach to this domain.
  
  \item Through the simulation experiments, this study demonstrated that despite operating without a replay buffer, IRRL achieved performance comparable to those of conventional replay buffer-based RL methods. Furthermore, it significantly outperformed existing incremental learning approaches.
  
  \item IRRL was validated through the real-world experiments, revealing that it can enable online adaptation to unseen environments and confirming its viability as a practical method for on-device learning in the real-world.

\end{itemize}

\section{RELATED WORKS}

\subsection{Social Navigation Using RL}
DRL has been widely applied to social navigation. Early works, such as the study reported in \cite{c26}, incorporated social awareness into collision avoidance. Because modeling human–robot interactions is essential, subsequent studies developed more sophisticated representations. For example, an attention-based pooling mechanism was established to evaluate the collective importance of neighboring humans \cite{c3}. Recently, graph neural networks (GNNs) have been heavily adopted to capture the complex relationships between robots and humans \cite{c4}–\cite{c29}, with some methods further integrating human trajectory predictions into the navigation policy \cite{c6}, \cite{c29}. Building upon these advancements, the proposed approach also integrates GNNs within an RL framework. However, a major limitation of these existing methods is their reliance on simulation-based training. Given the inherent unpredictability of human movement, policies trained solely in simulation often fail to generalize. Thus, real-world learning is essential, motivating us to establish a method tailored specifically for physical environments.

\subsection{Real-World RL}
Addressing the practical constraints of real-world RL has been the focus of several recent studies. For example, some approaches utilize reset policies to enable autonomous, unsupervised learning \cite{c30}, \cite{c27}, whereas others explore asynchronous RL \cite{c7}. To overcome the limitations of resource-constrained onboard computers, some frameworks distribute computational tasks to high-performance remote servers \cite{c8}. However, relying on continuous network connectivity introduces communication latency and vulnerability to network dropouts. Therefore, this study aim to ensure that the entire learning process can be executed locally on an edge device. To achieve this low-cost training, incremental learning is adopted. Incremental learning significantly reduces resource requirments by eliminating the large-scale replay buffers and computationally expensive batch updates. Moreover, training exclusively on the most recent data enables the system to adapt more rapidly to highly dynamic environments, such as crowded spaces, compared to methods that rely on potentially outdated experiences \cite{c10}. Specifically, the proposed framework builds upon Action Value Gradient (AVG) \cite{c9}, an entropy-maximizing DRL approach that mitigates the inherent instability of incremental learning through normalization and scaling.

\subsection{Efficient RL}
Training an RL policy from scratch requires vast amounts of interaction data, making it prohibitively time-consuming for real-world applications. Consequently, various methods have been proposed to accelerate learning. One prominent approach is residual RL \cite{c28}, \cite{c1}, which combines a predefined base policy with a learned RL policy. Other strategies include dynamically selecting between an imitation and RL policy based on action values \cite{c11}, learning a Gaussian policy in which the mean is optimized via RL while the variance is estimated from prior demonstrations \cite{c12}, and using the likelihood of an imitation policy as a regularization term during RL training \cite{c13}. In this study, the residual RL framework is employed. This approach is selected because its structural simplicity and minimal computational overhead make it exceptionally well-suited for real-world training on resource-constrained edge devices.

\section{PRELIMINARIES}

This study investigates the problem of a mobile robot navigating to a destination while avoiding pedestrians in a two-dimensional XY plane without static obstacles. In this situation, the robot operates in a body-fixed coordinate system, defined as a right-handed frame where the X-axis corresponds to the forward heading of the robot. The state of the robot at time step $t$ is defined by the goal coordinates, goal angle, and the velocity of the robot, all expressed in the robot's local coordinate system: $\mathbf{s}_t^r = [p_x^g, p_y^g, \theta^g, v_x, v_y]$. The state of the $i$-th pedestrian represents their position and relative velocity with respect to the robot's coordinate frame: $\mathbf{s}_t^i = [p_x^i, p_y^i, v_x^i, v_y^i]$. The aggregated state of all $n$ pedestrians in the environment is described by $\mathbf{s}_t^h = [\mathbf{s}_t^1, \mathbf{s}_t^2,..., \mathbf{s}_t^n]$. Additionally, because the robot is designed for omnidirectional movement, its action space is defined by the continuous velocity components in the X- and Y-directions: $\mathbf{a}_t = [v_x, v_y]$. In our RL framework, the reward function $r_t$ is formulated to encourage goal-reaching while penalizing collisions and uncomfortable proximity to pedestrians. It is defined as follows:

\begin{equation} 
r_t =
\begin{cases}
    -0.25 & \text{if } d_t < 0 \\
    (d_t-0.2)*0.125 & \text{else if } d_t < 0.2 \\
    1 & \text{else if } \mathbf{p}^r_t = \mathbf{p}^g \\
    0 & \text{otherwise}
\end{cases}\label{eq:reward}
\end{equation}

where $d_t$ is the minimum separation distance between the robot and any pedestrian, $\mathbf{p}_t^r$ is the current position of the robot, and $\mathbf{p}^g$ is the goal position of the robot.

\begin{figure*}[htb]
  \vspace*{1mm}
  \centering
  \includegraphics[width=0.93\textwidth]{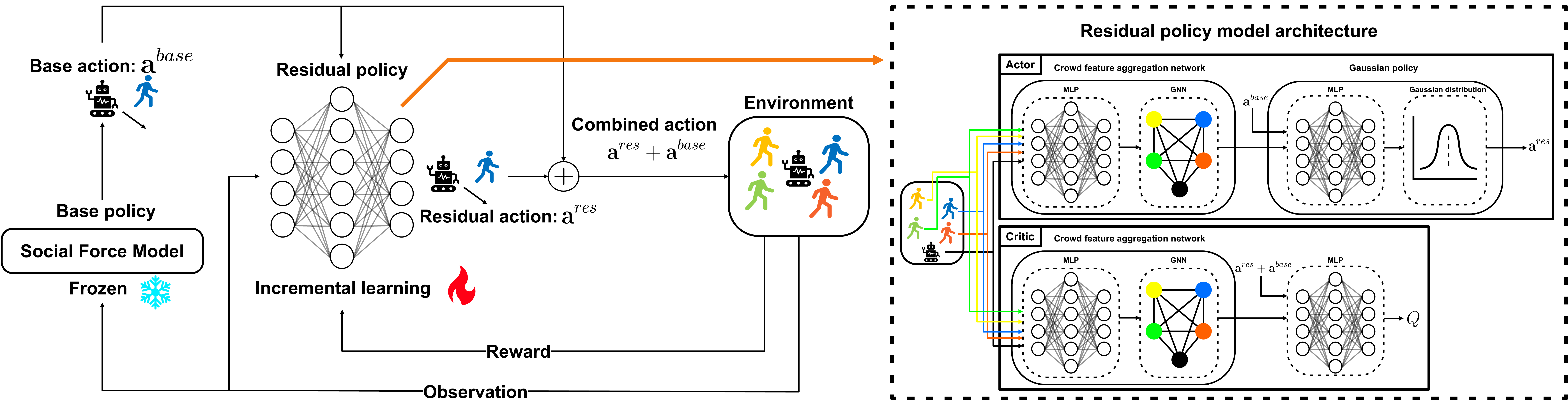}
  \caption{Illustration of IRRL framework and residual policy architecture for incremental learning. IRRL utilizes a residual RL framework that incrementally updates a residual policy alongside a frozen social force model as the base policy. The residual policy is trained within an actor–critic architecture, where both networks employ the GNN to aggregate crowd features.
  }
  \label{IRRL_overview}
\end{figure*}

\section{APPROACH}

This section details the learning methodology and model architecture of the proposed IRRL framework. An overview of IRRL is illustrated on the left side of Figure \ref{IRRL_overview}. IRRL framework combines a predefined base policy with a learned residual RL policy to enhance learning efficiency. Social force model (SFM) \cite{c14} is employed as the base policy in this study. SFM is an artificial potential field method that models pedestrian dynamics based on virtual attractive and repulsive forces. At each time step, the residual policy undergoes incremental learning, and the final executed action is the summation of the actions generated by the base and residual policies.

\subsection{Model Architecture of the Residual Policy}

The right side of Figure \ref{IRRL_overview} depicts the architecture of the residual policy, which employs an actor–critic framework. Both the actor and critic networks utilize the GNN, specifically GATv2 \cite{c15}, to aggregate state information from the robot and humans. Incorporating the GNN allows the model to flexibly handle environments with a variable number of observed pedestrians. The actor network outputs a Gaussian policy, generating an action distribution based on the aggregated crowd features and the action of the base policy. Meanwhile, the critic network evaluates the action-value (Q-value) by taking the aggregated crowd features and the sum of base and residual actions as inputs.

\subsection{Training the Residual Policy}

Algorithm \ref{residual_policy} outlines the training procedure for the residual policy. To stabilize the inherently unstable incremental learning process based on AVG, the policy employs entropy-maximizing RL integrated with the normalization and scaling techniques. For normalization, the penultimate normalization proposed in \cite{c17} is applied to both the actor and critic networks. This technique reduces variance in RL by dividing the feature vector of the second-to-last layer, $\psi_{\theta}(\mathbf{S})$, by its L2 norm: $\hat{\psi}_\theta(\mathbf{S}) = \frac{\psi_\theta(\mathbf{S})}{\| \psi_\theta(\mathbf{S}) \|_2}$. Furthermore, the temporal difference (TD) error is scaled using the method proposed in \cite{c18}, which divides the raw TD error by an online approximation of its standard deviation. Algorithm \ref{ScaleTDError} details this TD error scaling process, and Algorithm \ref{Normalize} presents the online method used to estimate the sample mean and variance. The temperature parameter, which regulates policy entropy, is automatically adjusted following the method in \cite{c16}. Because IRRL is designed for incremental learning, it strictly operates without a replay buffer or batch updates; instead, the model is updated solely using the latest transition data at each time step. Finally, to maintain architectural simplicity and reduce computational overhead, target networks and double Q-learning are explicitly excluded from the framework.

\vspace{-1mm}
\begin{algorithm}[htb]
    \caption{Training algorithm for the residual policy}
    \label{residual_policy}
    \begin{algorithmic}
    \STATE \textbf{Require:} Base policy $\pi_b$
    \STATE \textbf{Initialize:} $\gamma, H, \alpha$
    \STATE $\theta,\phi$ with penultimate normalization
    \STATE $\mathbf{n}_{\delta} = [0,0,0],\bm{\upmu}_{\delta} = [0,0,0],\bm{\bar{\upmu}}_{\delta} = [0,0,0]$
    \FOR{$n=0,...,N-1$ episodes}
        \STATE Initialize $\mathbf{s}
        _0, G = 0$
        \WHILE{s is not terminal}
            \STATE $\mathbf{a}_t^{base} = \pi_b(\mathbf{s}_t)$ and $\mathbf{a}_t^{res} \sim \pi_{\theta}(\mathbf{s}_t)$
            \STATE $\mathbf{u}_t= \mathbf{a}_t^{base} + \mathbf{a}_t^{res}$ 
            \STATE Take action $\mathbf{u}_t$ ,observe $\mathbf{s}_{t+1}$,$r_t$
            \STATE $G = G+r_t$
            \IF{$\mathbf{s}_{t+1}$ is terminal}
                \STATE \scalebox{0.985}{$\sigma_{\delta},\mathbf{n}_{\delta},\bm{\upmu}_{\delta},\bm{\bar{\upmu}}_{\delta} \gets$ScaleTDError$(r_t,0,G,\mathbf{n}_{\delta},\bm{\upmu}_{\delta},\bm{\bar{\upmu}}_{\delta})$} 
            \ELSE
                \STATE $\sigma_{\delta},\mathbf{n}_{\delta},\bm{\upmu}_{\delta},\bm{\bar{\upmu}}_{\delta} \gets $ScaleTDError$(r_t,\gamma,\emptyset,\mathbf{n}_{\delta},\bm{\upmu}_{\delta},\bm{\bar{\upmu}}_{\delta})$ 
            \ENDIF
            \STATE $\mathbf{a}_{t+1}^{base} = \pi_b(\mathbf{s}_{t+1})$ and $\mathbf{a}_{t+1}^{res} \sim \pi_{\theta}(\mathbf{s}_{t+1})$
            \STATE $\mathbf{u}_{t+1}= \mathbf{a}_{t+1}^{base} + \mathbf{a}_{t+1}^{res}$ 
            \STATE $\delta=r_t+\gamma(Q_\phi(\mathbf{s}_{t+1},\mathbf{u}_{t+1})-n\log \pi_\theta(\mathbf{a}_{t+1}^{res}|\mathbf{s}_{t+1}))-Q_\phi(\mathbf{s}_t,\mathbf{u}_t)$
            \STATE Update actor by minimizing
            \STATE \qquad$-Q_\phi(\mathbf{s}_t,\mathbf{u}_t)+\alpha\log\pi_\theta(\mathbf{a}_t^{res}|\mathbf{s}_t)$
            \STATE Update critic by minimizing
            \STATE \qquad$\delta/\sigma_{\delta}$
            \STATE Adjust temperature by minimizing
            \STATE \qquad$-\alpha\log\pi_\theta(\mathbf{a}_t^{res}|\mathbf{s}_t)-\alpha H$
            
        \ENDWHILE
    \ENDFOR
	\end{algorithmic}
\end{algorithm}

\begin{figure}[t]
    \vspace*{0.2mm} 
\end{figure}

\begin{algorithm}[htb]
\caption{ScaleTDError}
\label{ScaleTDError}
\begin{algorithmic}
    \STATE \textbf{Input} $R, \gamma, G, \mathbf{n}_\delta, \bm{\upmu}_\delta, \bm{\bar{\upmu}}_{\delta}$
    \STATE $n_R, n_\gamma, n_G \leftarrow \mathbf{n}_\delta; \quad \mu_R, \mu_\gamma, \mu_G \leftarrow \bm{\upmu}_\delta$
    \STATE $\bar{\mu}_R, \bar{\mu}_\gamma, \bar{\mu}_G \leftarrow \bm{\bar{\upmu}}_\delta$ 
    \STATE $n_R, \mu_R, \bar{\mu}_R, \sigma_R \leftarrow \text{Normalize}(R, n_R, \mu_R, \bar{\mu}_R)$
    \STATE $n_\gamma, \mu_\gamma, \bar{\mu}_\gamma, \sigma_\gamma \leftarrow \text{Normalize}(\gamma, n_\gamma, \mu_\gamma, \bar{\mu}_\gamma)$
    \IF{$G$ is not $\emptyset$}
        \STATE $n_G, \mu_G, \bar{\mu}_G, \_ \leftarrow \text{Normalize}(G^2, n_G, \mu_G, \bar{\mu}_G)$
    \ENDIF
    \IF{$n_G > 1$}
        \STATE $\sigma_\delta = \sqrt{\sigma_R + \mu_G \sigma_\gamma}$
    \ELSE
        \STATE $\sigma_\delta = 1$
    \ENDIF
    
    \STATE $\mathbf{n}_\delta = [n_R, n_\gamma, n_G]; \quad \bm{\upmu}_\delta = [\mu_R, \mu_\gamma, \mu_G]$
    \STATE $\bm{\bar{\upmu}}_\delta = [\bar{\mu}_R, \bar{\mu}_\gamma, \bar{\mu}_G]$
    
    \RETURN $\sigma_\delta, \mathbf{n}_\delta, \bm{\upmu}_\delta, \bm{\bar{\upmu}}_\delta$
\end{algorithmic}
\end{algorithm}

\begin{algorithm}[htb]
\caption{Normalize}
\label{Normalize}
\begin{algorithmic}
    \STATE \textbf{Input} $x, n, \mu, \bar{\mu}$
    
    \STATE $n = n + 1$
    \STATE $\delta = x - \mu$
    \STATE $\mu = \mu + \delta/n$
    \STATE $\delta_2 = x - \mu$
    \STATE $\bar{\mu} = \bar{\mu} + \delta \cdot \delta_2$
    \STATE $\sigma = \bar{\mu}/n$
    \RETURN $n, \mu, \bar{\mu}, \sigma$
\end{algorithmic}
\end{algorithm}

\section{SIMULATION EXPERIMENTS}

As described in this section, the simulation experiments were conducted to verify the performance of IRRL. The proposed method was evaluated by comparing it against three distinct categories of approaches: the base policy (SFM), RL methods utilizing a replay buffer, and methods that perform incremental learning without a replay buffer.

\subsection{Simulation Environment and Settings}

CrowdNav \cite{c3} was utilized as the simulation environment. The experiments were conducted in a circle-crossing scenario featuring five pedestrians. To evaluate adaptability to different pedestrian dynamics, scenarios in which pedestrians were controlled by either SFM or ORCA \cite{c19} were compared. \par
Each model was trained for 100,000 episodes, and the final evaluation was performed over 500 episodes using models trained across five different random seeds. The evaluation metrics included the success rate, collision rate, average execution time, and average return.\\
The following baselines were evaluated for the comparative analysis:
\begin{itemize}
    \item \textbf{Base policy:} SFM policy.
    \item \textbf{Replay buffer-based RL:} SAC \cite{c16}, \cite{c20}, TD3 \cite{c21}, and PPO \cite{c22} applied within the proposed residual RL framework.
    \item \textbf{Incremental RL (without a replay buffer):} Stream AC($\uplambda$) \cite{c10}, SAC-1, and TD3-1 \cite{c9}, also applied within the proposed residual RL framework.
\end{itemize}

\subsection{Performance Evaluation}

The numerical results of the evaluation episodes are presented in Table \ref{simulation_experiment}. The values represent the average across all seeds, with the $\pm$ symbol indicating the standard deviation.\par
\begin{figure}[htb]
  \centering
  \includegraphics[width=0.8\linewidth]{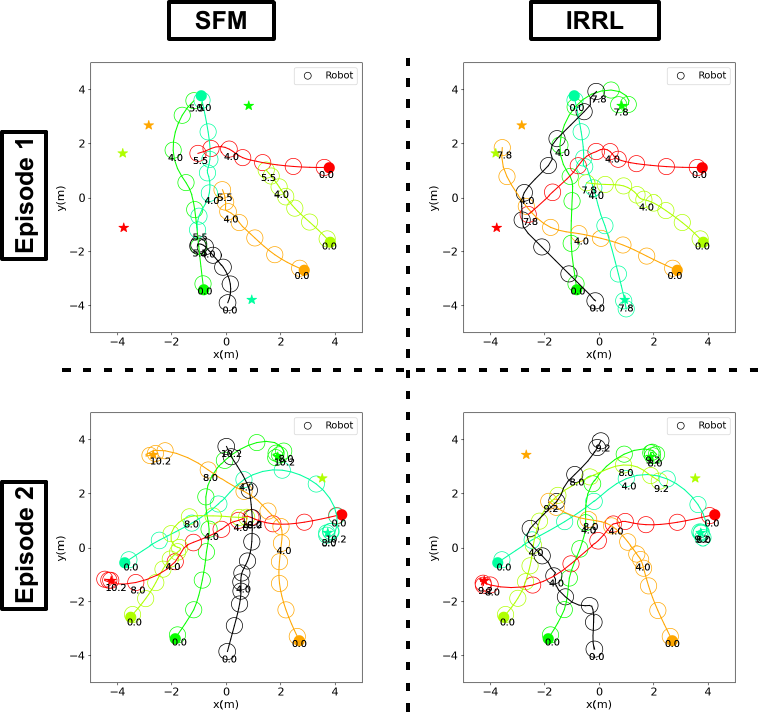}
  \caption{Navigation trajectories for each methods across two different evaluation episodes. The black lines represent the trajectories of the robot, whereas the colored lines denote individual pedestrian trajectories.
  }
  \label{traj_compare}
\end{figure}
\begin{figure}[htb]
  \centering
  \includegraphics[width=\linewidth]{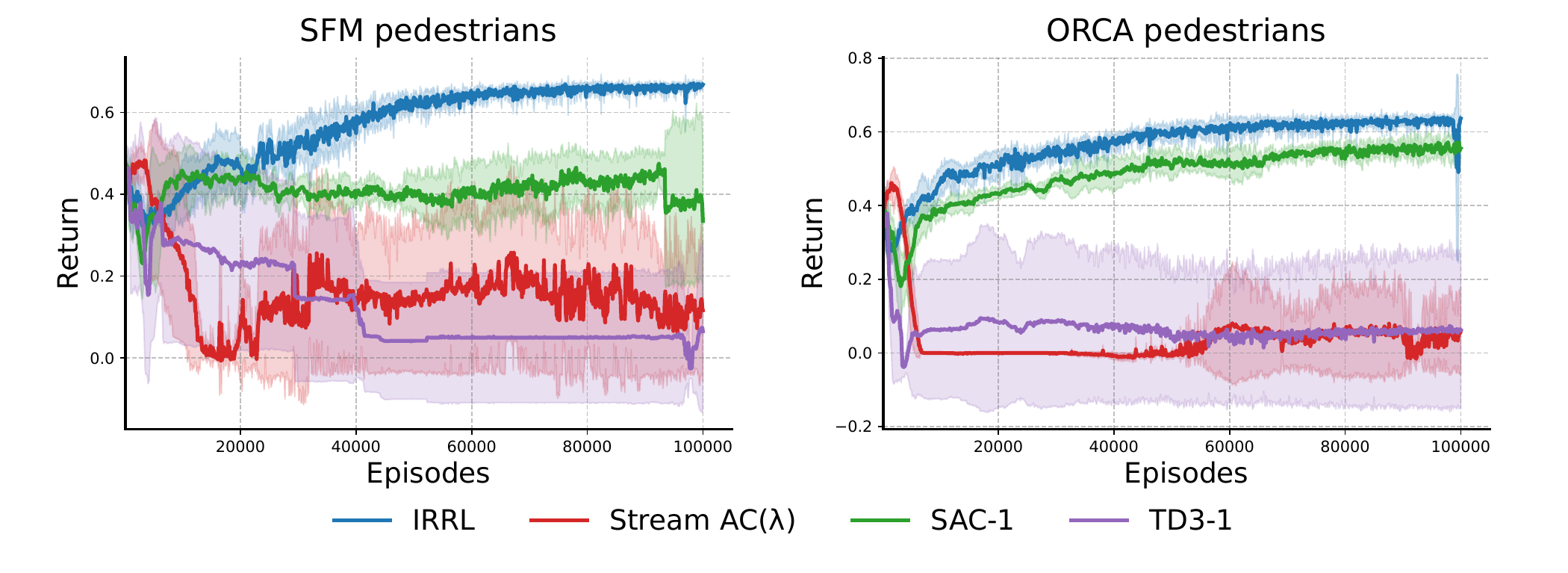}
  \caption{Learning curves of the return for methods trained via incremental learning (left: SFM pedestrians, right: ORCA pedestrians).}
  \label{Incremental plot}
\end{figure}

As indicated in Table \ref{simulation_experiment}, IRRL demonstrated improved performance across all metrics compared with SFM when pedestrians followed SFM. In the ORCA scenario, although its execution time was slightly longer, IRRL exhibited significant improvements in all other metrics. Figure \ref{traj_compare} illustrates the navigation trajectories of both IRRL and SFM across two sample episodes. The visualizations confirmed that IRRL successfully navigated the environment even in instances where SFM failed. Notably, in situations where SFM caused delays by passively waiting for pedestrians to pass, IRRL achieved earlier arrival by discovering more the efficient trajectory.
Compared with replay buffer-based methods, IRRL achieved competitive performance despite relying solely on incremental updates without a replay buffer.
Furthermore, among the incremental learning baselines, IRRL consistently delivered the best performance across all metrics and exhibited the smallest standard deviation across random seeds, highlighting its stability. Figure \ref{Incremental plot} shows the learning curves of the return for each incremental learning methods in both pedestrian scenarios, where each data point represents the average return of 100 evaluation episodes conducted every 100 training episodes. IRRL achieved the fastest convergence and highest overall return. Although SAC-1 exhibited some improvement in the ORCA scenario, it failed to exhibit meaningful performance gains in the SFM scenario. Moreover, depending on the seed, both Stream AC($\uplambda$) and TD3-1 experienced catastrophic failures, where the return droped to zero during training. By contrast, IRRL consistently improved the return and avoided such catastrophic failures in both dynamic environments.

\begin{figure*}[t]
    \vspace*{-2.5mm} 
\end{figure*}

\begin{table*}[htb]
    \centering
    \caption{Numerical comparison in the circle-crossing scenario with five pedestrians}
    \label{simulation_experiment}
    \resizebox{\linewidth}{!}{
    \begin{tabular}{lcccccccc}
        \toprule
        & \multicolumn{4}{c}{SFM pedestrians} & \multicolumn{4}{c}{ORCA pedestrians} \\
        \cmidrule(lr){2-5} \cmidrule(lr){6-9}
        Methods & Success[\%]$\uparrow$ & Collision[\%]$\downarrow$ & Exec. time[s]$\downarrow$ & Return$\uparrow$ & Success[\%]$\uparrow$ & Collision[\%]$\downarrow$ & Exec. time[s]$\downarrow$ & Return$\uparrow$\\ 
        \midrule
        SFM (base policy) & 0.806 $\pm$ 0.000 & 0.194 $\pm$ 0.000 & 10.00 $\pm$ 0.00 & 0.466 $\pm$ 0.000 & 0.786 $\pm$ 0.000 & 0.214 $\pm$ 0.000 & 8.71 $\pm$ 0.00 & 0.473 $\pm$ 0.000 \\ 

        \hdashline

        SAC (w/ buffer) & 0.999 $\pm$ 0.002 & 0.001 $\pm$ 0.002 & 7.79 $\pm$ 0.01 & 0.690 $\pm$ 0.001 & 0.998 $\pm$ 0.001 & 0.002 $\pm$ 0.001 & 8.11 $\pm$ 0.06 & 0.675 $\pm$ 0.001\\

        TD3 (w/ buffer) & 1.000 $\pm$ 0.000 & 0.000 $\pm$ 0.000 & 7.78 $\pm$ 0.01 & 0.691 $\pm$ 0.000 & 0.993 $\pm$ 0.003 & 0.007 $\pm$ 0.003 & 8.23 $\pm$ 0.15 & 0.668 $\pm$ 0.005\\

        PPO (w/ buffer) & 0.966 $\pm$ 0.006 & 0.034 $\pm$ 0.006 & 8.10 $\pm$ 0.08 & 0.651 $\pm$ 0.007 & 0.985 $\pm$ 0.003 & 0.014 $\pm$ 0.003 & 8.73 $\pm$ 0.11 & 0.645 $\pm$ 0.003\\

        \hdashline

        Stream AC($\uplambda$) (w/o buffer) & 0.911 $\pm$ 0.041 & 0.086 $\pm$ 0.041 & 12.30 $\pm$ 1.04 & 0.502 $\pm$ 0.013 & 0.906 $\pm$ 0.084 & 0.093 $\pm$ 0.085 & 14.11 $\pm$ 1.73 & 0.455 $\pm$ 0.038\\

        SAC-1 (w/o buffer) & 0.825 $\pm$ 0.092 & 0.169 $\pm$ 0.085 & 9.71 $\pm$ 0.83 & 0.485 $\pm$ 0.064 & 0.914 $\pm$ 0.069 & 0.084 $\pm$ 0.070 & 9.56 $\pm$ 0.97 & 0.554 $\pm$ 0.040\\

        TD3-1 (w/o buffer) & 0.873 $\pm$ 0.049 & 0.118 $\pm$ 0.061 & 10.21 $\pm$ 2.31 & 0.519 $\pm$ 0.039 & 0.910 $\pm$ 0.032 & 0.058 $\pm$ 0.037 & 15.56 $\pm$ 3.48 & 0.445 $\pm$ 0.053\\

        IRRL (w/o buffer) & 0.988 $\pm$ 0.003 & 0.012 $\pm$ 0.003 & 8.25 $\pm$ 0.12 & 0.666 $\pm$ 0.006 & 0.984 $\pm$ 0.006 & 0.015 $\pm$ 0.006 & 9.17 $\pm$ 0.18 & 0.635 $\pm$ 0.006\\

        \bottomrule
    \end{tabular}
    }
\end{table*}

\subsection{Effect of Residual RL}

\begin{figure}[htb]
  \centering
  \includegraphics[width=\linewidth]{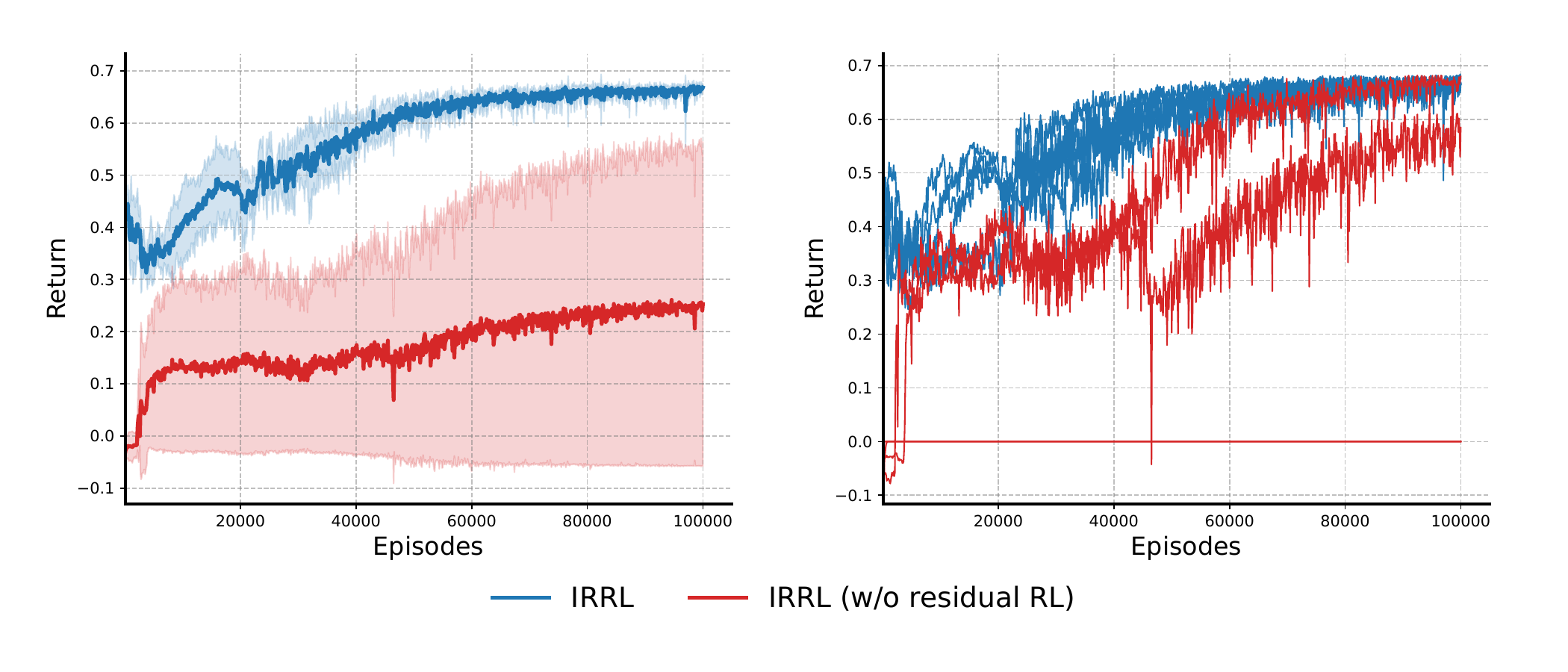}
  \caption{Learning curves of the return for IRRL and the ablated IRRL without residual RL. The left graph shows the mean learning curve across all random seeds, with the shaded region representing the standard deviation. The right graph shows the individual training runs for each seed.}
  \label{RRL eval plot}
\end{figure}

To evaluate the direct contribution of the residual RL component, an ablation study was conducted comparing IRRL with a variant trained from scratch (that is, without residual RL). Figure \ref{RRL eval plot} illustrates the learning curves of the return over 100,000 episodes across five random seeds in the SFM pedestrian environment.
The results clearly demonstrated that incorporating residual RL allowed IRRL to achieve faster and more consistent performance gains. Without the residual component, substantial variance was observed among the seeds, with some instances entirely failing to improve the return. These findings confirm that residual RL significantly enhances both the efficiency and stability of the incremental learning process.

\section{REAL-WORLD EXPERIMENTS}

\begin{figure}[htb]
  \centering
  \includegraphics[width=\linewidth]{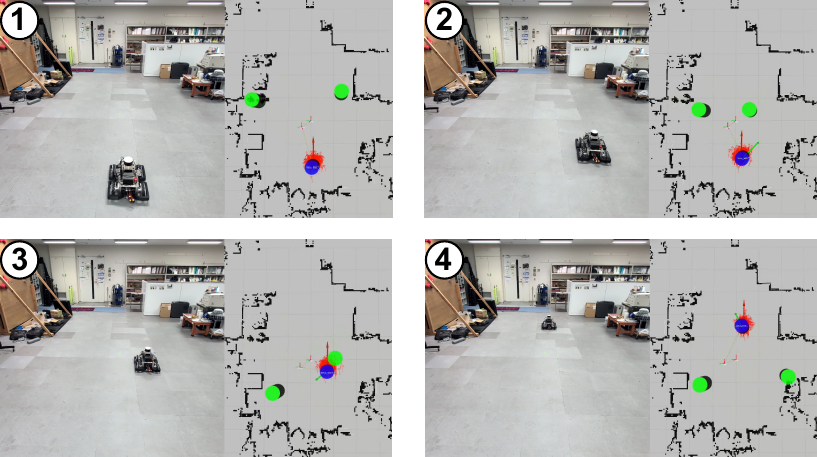}
  \caption{Scenes from the hybrid environment. The left side shows the physical robot operating in the physical environment, whereas the right side displays the corresponding RViz2 visualization. The blue cylinder represents the physical robot, and the green cylinders represet virtual pedestrians.}
  \label{Scenes of the real-world experiment}
\end{figure}

The real-world learning capabilities of the proposed IRRL framework were evaluated. The experimental scenario was structured to test the policy's adaptability to unseen dynamics. First, the robot was pre-trained for 100,000 episodes in a simulated circle-crossing environment featuring two cooperative pedestrians controlled by SFM who were aware of the position of the robot. 
Next, the robot was deployed in a physical environment containing two uncooperative pedestrians who were unaware of the position of the robot. 
The real-world experiment began with an initial evaluation of 50 episodes with virtual pedestrians and 30 episodes with physical pedestrians in a circle-crossing scenario. Subsequently, 100 episodes of online additional learning were conducted in this environment using IRRL.
Because maintaining continuous physical interactions with physical pedestrians for extended periods is highly impractical, the real-world RL framework that simulates virtual pedestrians within the physical space was developed and used for learning and evaluation. Figure \ref{Scenes of the real-world experiment} illustrates this hybrid environment. After completing the 100 training episodes, a final test was conducted in the virtual and physical pedestrians environment to verify the perfomance of the adapted policy.

\subsection{Hardware and Software Design}

The robotic platform was based on a Mecanum-wheeled rover equipped with a 3D-LiDAR (AR-XT-32-a) and an onboard computer (Jetson AGX Orin) that handles all computational processing locally. The software stack was developed using ROS 2 Humble. For physical pedestrians detection, DR-SPAAM \cite{c23} was employed, utilizing a pre-trained model optimized on the dataset provided in \cite{c31}. Because the baseline DR-SPAAM model frequently misclassifies static obstacles as pedestrians, an additional filtering step was implemented to remove static objects from the detection candidates. Furthermore, localization was achieved using monte carlo localization (MCL) with expansion resetting (emcl) \cite{c24}, \cite{c25}.

\subsection{Performance Evaluation}

\begin{figure}[htb]
  \vspace*{2mm}
  \centering
  \includegraphics[width=\columnwidth]{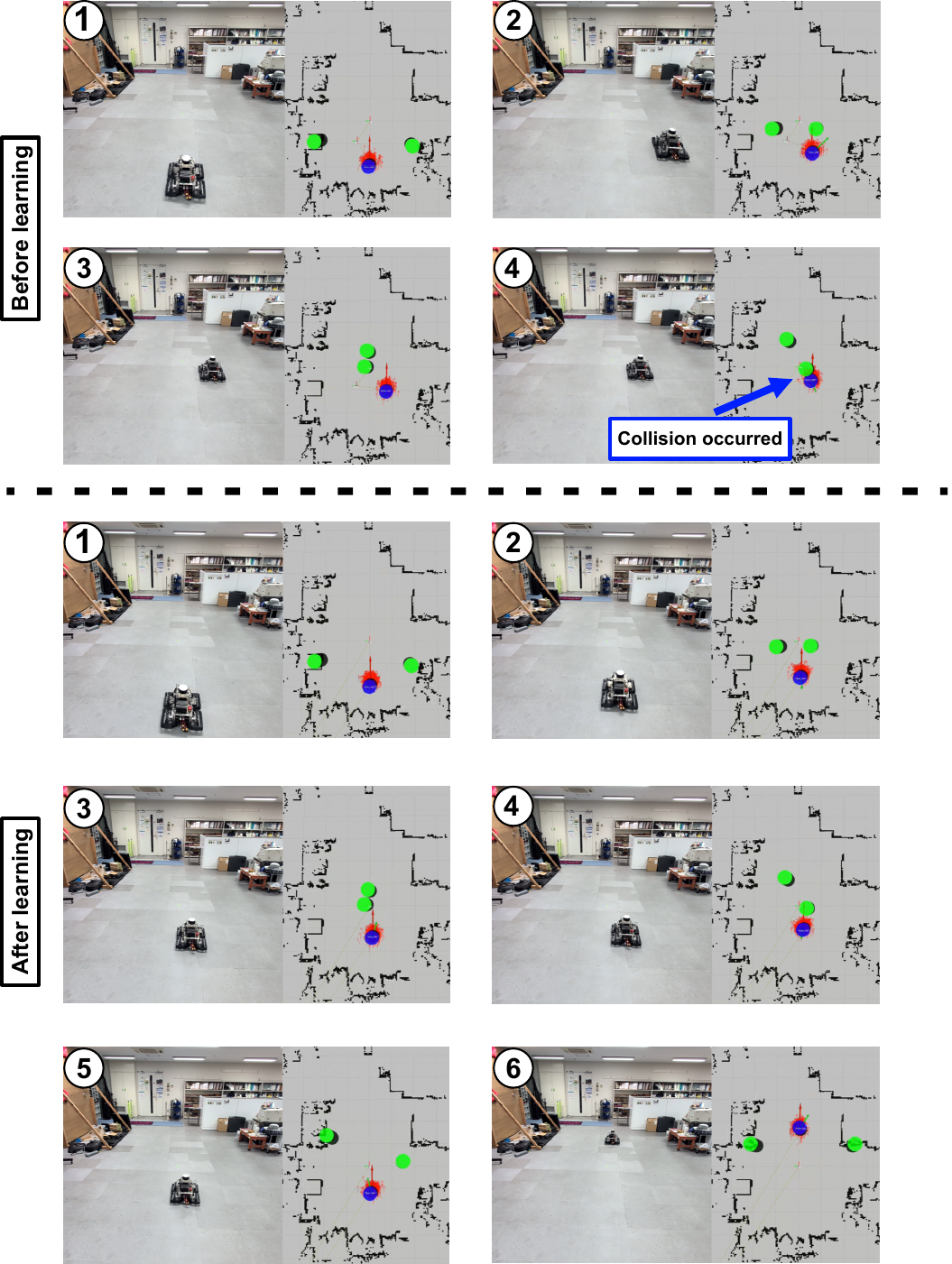}
  \caption{Comparison of the robot navigation behavior before and after the online learning phase within the hybrid environment.}
  \label{mix_compare}
\end{figure}

\begin{figure}[htb]
  \vspace*{2mm}
  \centering
  \includegraphics[width=\columnwidth]{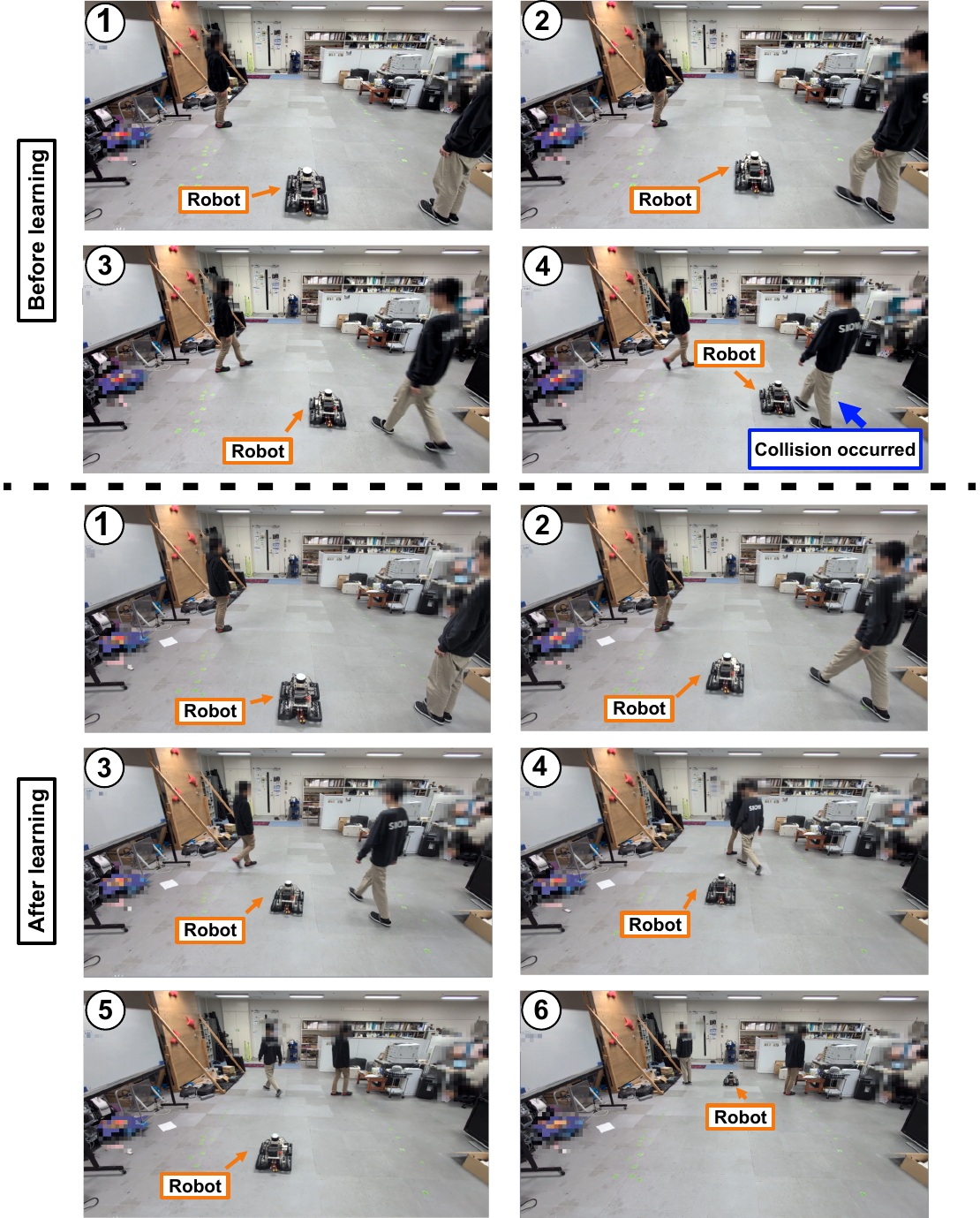}
  \caption{Comparison of the robot navigation behavior before and after the online learning phase within the physical pedestrians environment.}
  \label{real_compare}
\end{figure}

Table \ref{real_world_result} presents a numerical comparison for both virtual and physical pedestrians before and after 100 learning episodes in the hybrid environment. As shown in table \ref{real_world_result}, performance improvements were confirmed across all metrics, with the exception of execution time, in both the virtual and physical pedestrians scenarios. The prolonged execution time can be attributed to the difference in pedestrian behavior. When interacting with cooperative pedestrians, the robot can take a more efficient route and pass in front of them, as the pedestrians will also take action to mutually avoid the robot. However, uncooperative pedestrians do not pay attention to the robot. Consequently, attempting to cross their path frequently leads to collisions. To avoid such collisions, the robot must wait for these uncooperative pedestrians to pass, which we consider to be the primary reason for the increased execution time. Furthermore, Figures \ref{mix_compare} and \ref{real_compare} compare the robot navigation behavior before and after learning in the virtual and physical pedestrians test scenario, respectively. In both cases, although the policy before learning resulted in a collision when attempting to cross in front of a pedestrian, the policy after learning successfully avoided collisions by waiting for the pedestrian to pass. These results demonstrate that conducting real-world learning using IRRL enables the acquisition of a policy effectively adapted to the given environment.

\begin{table}[htb]
    \centering
    \caption{Numerical comparison in the real-world experiments}
    \label{real_world_result}
    \resizebox{\linewidth}{!}{%
    \begin{tabular}{lccccc}
        \toprule
        Environment & Method & Success [\%]$\uparrow$ & Collision [\%]$\downarrow$ & Exec. time [s]$\downarrow$ & Return$\uparrow$ \\ \hline
        virtual pedestrians & Before & 0.520 & 0.440 & \textbf{6.56} & 0.276\\
        50 trials & After & \textbf{0.820} & \textbf{0.180} & 7.30 & \textbf{0.537} \\ 
        \hdashline
        physical pedestrians & Before & 0.400 & 0.600 & \textbf{8.75} & 0.129\\
        30 trials & After & \textbf{0.600} & \textbf{0.400} & 9.01 & \textbf{0.299}\\ 
        \bottomrule
    \end{tabular}
    }
\end{table}

\section{CONCLUSION}









In this study, we proposed IRRL, a novel RL framework toward real-world learning for social navigation that integrates the lightweight nature of incremental learning with the efficiency of residual RL. Through the simulation experiments, this study demonstrated that despite operating without a replay buffer or batch updates, IRRL achieved a performance comparable to those of conventional replay buffer-based methods while significantly outperforming existing incremental learning approaches. Furthermore, we confirmed that the policy successfully improved its performance through the real-world learning. Crucially, the entire real-world learning process can be successfully executed on a resource-constrained edge device during operation. In the future, we will focus on mobile robots capable of adapting to pedestrian environments that evolve over time by pursuing continuous, lifelong learning.

\section*{ACKNOWLEDGMENT}

This work was partially supported by JSPS KAKENHI Grant Number JP26K21355.


\begin{thebibliography}{99}

\bibitem{c2} Ibrahim Khalil Kabir, and Muhammad Faizan Mysorewala, "Socially aware navigation for mobile robots: a survey on deep reinforcement learning approaches," {\it Applied Intelligence}, vo. 56, no. 1, pp.38, 2026.

\bibitem{c26} Yu Fan Chen, Michael Everett, and Jonathan P. How, "Socially aware motion planning with deep reinforcement learning," in {\it Proceedings of the IEEE/RSJ International Conference on Intelligent Robots and Systems (IROS)}, pp. 1343-1350, 2017.

\bibitem{c3} Changan Chen, Yuejiang Liu, Sven Kreiss, and Alexandre Alahi, "Crowd-robot interaction: Crowd-aware robot navigation with attention-based deep reinforcement learning," in {\it Proceedings of the IEEE International Conference on Robotics and Automation (ICRA)}, pp. 6015-6022, 2019.

\bibitem{c4} Yuying Chen, Congcong Liu, Bertram E. Shi, and Ming Liu, "Robot navigation in crowds by graph convolutional networks with attention learned from human gaze," {\it IEEE Robotics and Automation Letters}, vo. 5, no. 2, pp. 2754-2761, 2020.

\bibitem{c5} Xueyou Zhang, Wei Xi, Xian Guo, Yongchun Fang, Bin Wang, Wulong Liu, and Jianye Hao, "Relational navigation learning in continuous action space among crowds," in {\it Proceedings of the IEEE International Conference on Robotics and Automation (ICRA)}, pp. 3175-3181, 2021.

\bibitem{c6} Changan Chen, Sha Hu, Payam Nikdel, Greg Mori, and Manolis Savva, "Relational graph learning for crowd navigation," in {\it Proceedings of the IEEE/RSJ International Conference on Intelligent Robots and Systems (IROS)}, pp. 10007-10013, 2020.

\bibitem{c29} Shuijing Liu, Peixin Chang, Zhe Huang, Neeloy Chakraborty, Kaiwen Hong, Weihang Liang, D. Livingston McPherson, Junyi Geng, and Katherine Driggs-Campbell, "Intention Aware Robot Crowd Navigation with Attention-Based Interaction Graph," in {\it Proceedings of the IEEE International Conference on Robotics and Automation (ICRA)}, pp. 12015-12021, 2023.

\bibitem{c30} Laura Smith, J. Chase Kew, Xue Bin Peng, Sehoon Ha, Jie Tan, and Sergey Levine, "Legged Robots that Keep on Learning: Fine-Tuning Locomotion Policies in the Real World," in {\it Proceedings of the IEEE International Conference on Robotics and Automation (ICRA)}, pp. 1593-1599, 2022.

\bibitem{c27} Russell Mendonca, Emmanuel Panov, Bernadette Bucher, Jiuguang Wang, and Deepak Pathak, "Continuously Improving Mobile Manipulation with Autonomous Real-World RL," in {\it Proceedings of the Conference on Robot Learning (CoRL)}, pp. 5024-5219, 2024.

\bibitem{c9} Gautham Vasan, Mohamed Elsayed, Seyed Alireza Azimi, Jiamin He, Fahim Shahriar, Colin Bellinger, Martha White, and Rupam Mahmood, "Deep policy gradient methods without batch updates, target networks, or replay buffers," in {\it Advances in Neural Information Processing Systems (NeurIPS)}, pp. 845-891, 2024.

\bibitem{c10} Mohamed Elsayed, Gautham Vasan, and A Rupam Mahmood, "Streaming deep reinforcement learning finally works," {\it CoRR}, vol. abs/2410.14606, 2024.

\bibitem{c28} Tom Silver, Kelsey R. Allen, Josh Tenenbaum, and Leslie Pack Kaelbling, "Residual Policy Learning," {\it CoRR}, vol. abs/1812.06298, 2018.

\bibitem{c1} Tobias Johannink, Shikhar Bahl, Ashvin Nair, Jianlan Luo, Avinash Kumar, Matthias Loskyll, Juan Aparicio Ojea, Eugen Solowjow, and Sergey Levine, "Residual reinforcement learning for robot control,"  in {\it Proceedings of the IEEE International Conference on Robotics and Automation (ICRA)}, pp. 6023-6029, 2019.

\bibitem{c7} Yufeng Yuan, and A. Rupam Mahmood, "Asynchronous reinforcement learning for real-time control of physical robots," in {\it Proceedings of the IEEE International Conference on Robotics and Automation (ICRA)}, pp. 5546-5552, 2022.

\bibitem{c8} Yan Wang, Gautham Vasan, and A. Rupam Mahmood, "Real-time reinforcement learning for vision-based robotics utilizing local and remote computers," in {\it Proceedings of the IEEE International Conference on Robotics and Automation (ICRA)}, pp. 9435-9441, 2023.

\bibitem{c11} Hengyuan Hu, Suvir Mirchandani, and Dorsa Sadigh, "Imitation bootstrapped reinforcement learning," in {\it Proceedings of the Robotics: Science and Systems (RSS)}, 2024.

\bibitem{c12} Perry Dong, Alec M. Lessing, Annie S. Chen, and Chelsea Finn, "Reinforcement learning via implicit imitation guidance," {\it CoRR}, vol. abs/2506.07505, 2025.

\bibitem{c13} Chenran Li, Chen Tang, Haruki Nishimura, Jean Mercat, Masayoshi TOMIZUKA, and Wei Zhan, "Residual q-learning: Offline and online policy customization without value," in {\it Advances in Neural Information Processing Systems (NeurIPS)}, pp. 61857-61869, 2023.

\bibitem{c14} Dirk Helbing, and Péter Molnár, "Social force model for pedestrian dynamics," {\it Physical Review E}, pp. 4282-4286, 1995. 

\bibitem{c15} Shaked Brody, Uri Alon, and Eran Yahav, "How attentive are graph attention networks?," in {\it Proceedings of the International Conference on Learning Representations (ICLR)}, 2022.

\bibitem{c17} Johan Bjorck, Carla P. Gomes, and Kilian Q. Weinberger, "Is high variance unavoidable in rl? A case study in continuous control," in {\it Proceedings of the International Conference on Learning Representations (ICLR)}, 2022.

\bibitem{c18} Tom Schaul, Georg Ostrovski, Iurii Kemaev, and Diana Borsa, "Return-based scaling: Yet another normalisation trick for deep RL," {\it CoRR}, vol. abs/2105.05347, 2021.

\bibitem{c16} Tuomas Haarnoja, Aurick Zhou, Kristian Hartikainen, George Tucker, Sehoon Ha, Jie Tan, Vikash Kumar, Henry Zhu, Abhishek Gupta, Pieter Abbeel and Sergey Levine, "Soft actor-critic algorithms and applications," {\it CoRR}, vol. abs/1812.05905, 2018.

\bibitem{c19} Jur van den Berg, Stephen J. Guy, Ming C. Lin, and Dinesh Manocha, "Reciprocal n-body collision avoidance," in {\it Proceedings of the International Symposium of Robotics Research(ISRR)}, pp. 3-19, 2009.

\bibitem{c20} Tuomas Haarnoja, Aurick Zhou, Pieter Abbeel, and Sergey Levine, "Soft actor-critic: Off-policy maximum entropy deep reinforcement learning with a stochastic actor," in {\it Proceedings of the International Conference on Machine Learning (ICML)}, pp. 1856-1865, 2018.

\bibitem{c21} Scott Fujimoto, Herke van Hoof, and David Meger, "Addressing function approximation error in actor-critic methods," in {\it Proceedings of the International Conference on Machine Learning (ICML)}, pp. 1582-1591, 2018.

\bibitem{c22} John Schulman, Filip Wolski, Prafulla Dhariwal, Alec Radford, and Oleg Klimov, "Proximal policy optimization algorithms," {\it CoRR}, vol. abs/1707.06347, 2017.

\bibitem{c23} Dan Jia, Alexander Hermans, and Bastian Leibe, "DR-SPAAM: A Spatial-Attention and Auto-regressive Model for Person Detection in 2D Range Data," in {\it Proceedings of the IEEE/RSJ International Conference on Intelligent Robots and Systems (IROS)}, pp. 10270–10277, 2020.

\bibitem{c31} Fernando Amodeo, No{\'{e}} P{\'{e}}rez{-}Higueras, Luis Merino, and Fernando Caballero, "{FROG:} a new people detection dataset for knee-high 2D range finders," {\it Frontiers Robotics {AI}}, vo. 12, 2025.

\bibitem{c24} R. Ueda, "Syokai Kakuritsu Robotics (lecture note on probabilistic robotics)," Kodansya, 2019.

\bibitem{c25} R. Ueda, T. Arai, K. Sakamoto, T. Kikuchi, and S. Kamiya, "Expansion resetting for recovery from fatal error in monte carlo localization - comparison with sensor resetting methods," in {\it Proceedings of the IEEE/RSJ International Conference on Intelligent Robots and Systems (IROS)}, pp. 2481–2486, 2004


\end{thebibliography}
\end{document}